\documentclass{article}

\usepackage[preprint]{neurips_2024}

\usepackage[utf8]{inputenc} 
\usepackage[T1]{fontenc}    
\usepackage{hyperref}       
\usepackage{url}            
\usepackage{booktabs}       
\usepackage{amsfonts}       
\usepackage{nicefrac}       
\usepackage{microtype}      
\usepackage{xcolor}         
\usepackage{ulem}
\usepackage{graphicx}
\usepackage{array}

\usepackage{amsmath,amsfonts,amssymb,amsthm, bbm}
\usepackage{algorithm,amsmath}
\usepackage{algpseudocode}
\title{PG-Rainbow: Using Distributional Reinforcement Learning in Policy Gradient Methods}
\author{
    WooJae Jeon\thanks{jjjeon0223@snu.ac.kr.} \\
    Seoul National University\\
    \and 
    KangJun Lee\thanks{kglee9128@soongsil.ac.kr} \\
    Soongsil University
    \and 
    Jeewoo Lee\thanks{leejeewoo@berkeley.edu} \\
    UC Berkeley
}

\begin{document}

\maketitle

\begin{abstract}
This paper introduces PG-Rainbow, a novel algorithm that incorporates a distributional reinforcement learning framework with a policy gradient algorithm. Existing policy gradient methods are sample inefficient and rely on the mean of returns when calculating the state-action value function, neglecting the distributional nature of returns in reinforcement learning tasks. To address this issue, we use an Implicit Quantile Network that provides the quantile information of the distribution of rewards to the critic network of the Proximal Policy Optimization algorithm. We show empirical results that through the integration of reward distribution information into the policy network, the policy agent acquires enhanced capabilities to comprehensively evaluate the consequences of potential actions in a given state, facilitating more sophisticated and informed decision-making processes. We evaluate the performance of the proposed algorithm in the Atari-2600 game suite, simulated via the Arcade Learning Environment (ALE).
\end{abstract}

\section{Introduction}
In recent years, significant improvements have been made in deep reinforcement learning, with multiple algorithms showing state-of-the-art performance across diverse domains and environments (\citet{DBLP:journals/corr/abs-1912-06680}, \citet{dqnatari}). Amidst the various branches in deep reinforcement learning, the field of on-policy reinforcement learning has attracted great attention because it directly interacts with the environment during training, offering advantages in quicker training time and the ability to adapt to changing conditions in real-time, thereby holding promise for applications in dynamic and uncertain environments.

Though on-policy algorithms show great performance, it has the critical problem of sample inefficiency. The policy can only be optimized using experiences collected by the latest policy network, which leads to longer training time and higher variance. Though several approaches, such as constraining the policy to learn in small steps (\citet{DBLP:journals/corr/SchulmanLMJA15}) or using entropy regularization (\citet{DBLP:journals/corr/abs-1801-01290}), were suggested to mitigate this issue, on-policy algorithms are still more likely to converge to a local optima than off-policy algorithms. Also, current on-policy reinforcement learning ignores the distribution of rewards in training, which potentially limits its ability to capture the full spectrum of uncertainty and variability inherent in complex environments. 

To overcome these drawbacks, we propose a method that uses an off-policy distributional reinforcement learning algorithm that provides the policy network with information about the reward distribution. While distributional reinforcement learning has been integrated with various off-policy value function methods (\citet{rainbow}), to the best of our knowledge, the utilization of this approach in conjunction with on-policy algorithms remains relatively unexplored. By using distributional reinforcement learning, we alleviate the sample inefficiency inherent in on-policy algorithms by utilizing experiences that would have been discarded to train an off-policy algorithm, thereby enhancing data efficiency and potentially improving learning efficiency in reinforcement learning tasks.

In this paper we introduce a novel reinforcement learning algorithm called PG-Rainbow that distills information produced by distributional reinforcement learning \citep{DBLP:journals/corr/BellemareDM17} to an on-policy algorithm, PPO (\citet{DBLP:journals/corr/SchulmanWDRK17}). By feeding the reward distribution information into the policy network, we anticipate that the policy agent will take into account of state-action value distribution when making action selection decisions in any given state. 

The main contributions of our paper is as follows:
\begin{enumerate}
  \item We explore the field of integrating distributional reinforcement learning algorithm into on-policy reinforcement learning.
  \item We alleviate the sample inefficiency problem of on-policy algorithms by storing experiences collected to train an off-policy distributional reinforcement learning algorithm.
\end{enumerate}

\section{Related Work}
Consider an environment modeled as a Markov Decision Process (MDP), denoted $(\mathcal{S},\mathcal{A},R,P,\gamma)$, where $\mathcal{S}$ represents the state space, $\mathcal{A}$ represents the action space, $R:\mathcal{S}\rightarrow\mathbb{R}$ represents the rewards function, $P:\mathcal{S}\times\mathcal{A}\times\mathcal{S}\rightarrow\mathbb{R}$ is the transition probability function and $\gamma\in[0,1]$ a discount value. A policy $\pi(\cdot|s)$ is a function that maps an arbitrary state $s$ to a distribution of available actions. Denote $v_{\pi}(s)$ as the expected return of an agent being positioned in state $s$ and following $\pi$.

The goal of a reinforcement learning agent is to maximize its cumulative rewards attained over the course of an episode, where an episode is essentially a trajectory of states and actions that the agent went through until the agent has arrived at a terminal state. Hence, the agent is given the task of maximizing the return $\sum_{t=0}^{\infty}\gamma^{t}R(s_{t},a_{t})$, which is the discounted sum of all future rewards starting from the initial state $s_{0}$. For an agent following policy $\pi$, the action-value function is defined as $Q^{\pi}(s,a)=\mathbb{E}[\sum_{t=0}^{\infty}\gamma^{t}R(s_{t},a_{t})]$ and the Bellman equation is defined as: \[
Q^{\pi}(s,a)=\mathbb{E}[R(s,a)] + \gamma\mathbb{E}[Q^{\pi}(s',a')]
\]

To find an optimal policy $\pi^{*}$, \citet{Bellman:DynamicProgramming} introduced the Bellman optimality operator, where the optimal policy can be obtained by iteratively improving the action-value function (\citet{Watkins}). In essence, the policy is optimized by accurately approximating the state-action value function, and the policy selects the action that returns the highest action-value $\pi^{*}(s) = \arg\max_{a} Q^{\pi^{*}}(s,a)$.

\subsection{Policy Gradient Methods}
In policy gradient methods, the agent attempts to maximize the return by learning a parameterized policy $\pi_{\theta}$ that can select the optimal action given the current state. The parameter is updated using a gradient estimator $\hat{g}$, where $\alpha$ is a step size (\citet{Sutton1998}). 
\[\theta \leftarrow \theta + \alpha \hat{g}\]

The estimator is computed by differentiating the objective function $L(\theta)$. The objective function takes the following form, where $A_{t}$ is the advantage function at timestep $t$. \[
L(\theta)=\mathbb{E}[\log \pi_{\theta}(a_{t}\mid s_{t})A_{t}]
\]

However, policy gradient methods are more prone to irreversible huge gradient steps, which may lead to a terrible new policy $\pi_{new}$. Hence, Trust Region Policy Optimization (TRPO) suggests a constraint to the policy parameter update to prevent extensive changes to policy parameters (\citet{DBLP:journals/corr/SchulmanLMJA15}). Nevertheless, TRPO deals with second order optimizations, which adds complications.    

Proximal Policy Optimization (PPO) introduces the constraint by using a clip function that sets a upper and lower bound for the gradient step (\citet{DBLP:journals/corr/SchulmanWDRK17}). In PPO, the loss function is defined as: 
\[
L^{CLIP}(\theta)=\mathbb{E}_{t}\bigg[\min(r_{t}(\theta)A_{t},\text{clip}(r_{t}(\theta),1-\epsilon,1+\epsilon)) \bigg]
\] where $r_{t}(\theta)=\frac{\pi_{\theta}(a_{t}\mid s_{t})}{\pi_{\theta_{\text{old}}}(a_{t}\mid s_{t})}$ and $\epsilon$ is a hyperparameter to control the degree of clipping. Clipping the policy gradient has shown to significantly increase the stability and performance of the learned policy, with a much simpler implementation scheme. 

One limitation of PPO is that sharing features between the policy function and value function degrades the performance of the model. In the standard PPO model architecture, the policy function and value function share the same network structure and parameters, with the output head being the sole difference (Figure 1.). It was shown this leads to significant generalization problems, where the model cannot efficiently adapt to different but similar environments (\citet{DBLP:journals/corr/abs-2102-10330}).

Several approaches to improve the vanilla PPO algorithm have been introduced in the past by enhancing the value network of PPO. Phasic Policy Gradient (PPG) decouples the optimization process of the policy and value function by using two alternating phases in its training process. The first process called the policy phase trains the agent with the normal PPO algorithm. The auxiliary phase distills features from the value function into the policy network, to improve training in future policy phases (\citet{phasicpolicygradient}). Delayed-Critic Policy Gradient (DCPG) instead uses a single unified network architecture and addresses the memorization problem, which leads to generalization problems, of the value network by using stiffness analysis. In DCPG, the value network is trained with more data in the auxiliary phase and is optimized with a delay compared to the policy network (\citet{moon2023rethinking}). 

\subsection{Distributional Reinfocement Learning}
Distributional reinforcement learning differs from traditional reinforcement learning by considering the return distribution of an environment. Formal reinforcement learning assumes that the return generated by taking an action $a$ in state $s$ is the expected return. However, the reward distribution is rarely a normal distribution, and taking a simple expected value is not sufficient to capture the possible mutli-modality of the reward distribution (\citet{distrlbook}).

Distributional reinforcement learning incorporates the reward distribution by using a distributional Bellman equation defined as: \[
Z(s,a)  \overset{D}{=} R(s,a) + \gamma Z(S',A')
\] where the state-action value $Q(s,a)$ is replaced by value distribution $Z(s,a)$. The distributional Bellman optimality operator is then defined as: \[
\mathcal{T}Z(s,a) :\overset{D}{=} R(s,a) + \gamma Z(S',\underset{a'\in\mathcal{A}}{\arg\max}\mathbb{E}[Z(S',a')])
\]
\citet{DBLP:journals/corr/BellemareDM17} proved that the distributional Bellman operator for policy evaluation and control to be a contraction in the $p$-Wasserstein distance and proposed the C51 algorithm, where the reward distribution is projected over a fixed set of equidistant values. However, the C51 algorithm does not strictly follow the contraction theory, as it minimizes the KL divergence between the current and target distributions, instead of using the Wasserstein distance. 

To address the disparity between theory and practical applications, methodologies employing quantile regression have been developed. It is shown that using quantile regression in distributional reinforcement learning makes the projected distributional Bellman operator a contraction in the $\infty$-Wasserstein metric (\citet{qrdqn}).

The Implicit Quantile Network (IQN) follows the quantile regression approach, where it randomly samples the quantile values to learn an implicit representation of the return distribution. Let $F_{Z}^{-1}(\tau)$ be the quantile function at $\tau\in[0,1]$, where $\tau$ is a random sample from the base reward distribution. IQN models the state-action quantile function as a mapping from state-actions and samples from a base distribution to $Z_{\tau}(s,a)$. For two samples $\tau,\tau'\in U([0,1])$ and policy $\pi_{\beta}$, the sampled TD error at step $t$ is defined as: \[
\delta_{t}^{\tau,\tau'}=r_{t} + \gamma Z_{\tau'}(s_{t+1},\pi_{\beta}(s_{t+1}))-Z_{\tau}(s_t,a_t)
\]
The IQN loss function is then defined as: \[
\mathcal{L}(s_t,a_t,r_t,x_{t+1})=\frac{1}{N'}\sum_{i=1}^{N}\sum_{j=1}^{N'}\rho_{\tau_i}^{\kappa}\big(\delta_{t}^{\tau_i,\tau_{j}'} \big)
\] where $\rho_{\tau}^{\kappa} =\lvert\tau-\mathbbm{1}\{\delta_{ij}<0\}\rvert\frac{\mathcal{L}_{\kappa}(\delta_ij)}{\kappa}$. $N$ and $N'$ denote the respective number of i.i.d samples $\tau_i,\tau_{j}'\in U([0,1])$ used to estimate the loss. The quantile estimates are trained using the Huber quantile regression loss (\citet{american1972annals}), with threshold $\kappa$. \[
\mathcal{L}_{\kappa}(\delta_ij)=\begin{cases}
\frac{1}{2}\delta_{ij}^{2}&\text{if $\lvert\delta_{ij}\rvert\leq\kappa$}\\
\kappa(\lvert\delta_{ij}\rvert-\frac{1}{2}\kappa)&\text{otherwise}
\end{cases}
\]
The IQN model has been proved to be more robust and flexible in approximating the quantile values of a base distribution and has been the state-of-the-art algorithm for distributional reinforcement learning at the time of its introduction. It is also seamlessly integrated into existing deep learning architectures such as Double DQN (\citet{doubledqn}) or Dueling DQN (\citet{duelingdqn}), and was used in the Rainbow model (\citet{rainbow}), which achieved the state-of-the-art model for solving the Atari-2600 environment. 

\begin{figure}
  \centering
  \includegraphics[scale=0.3]{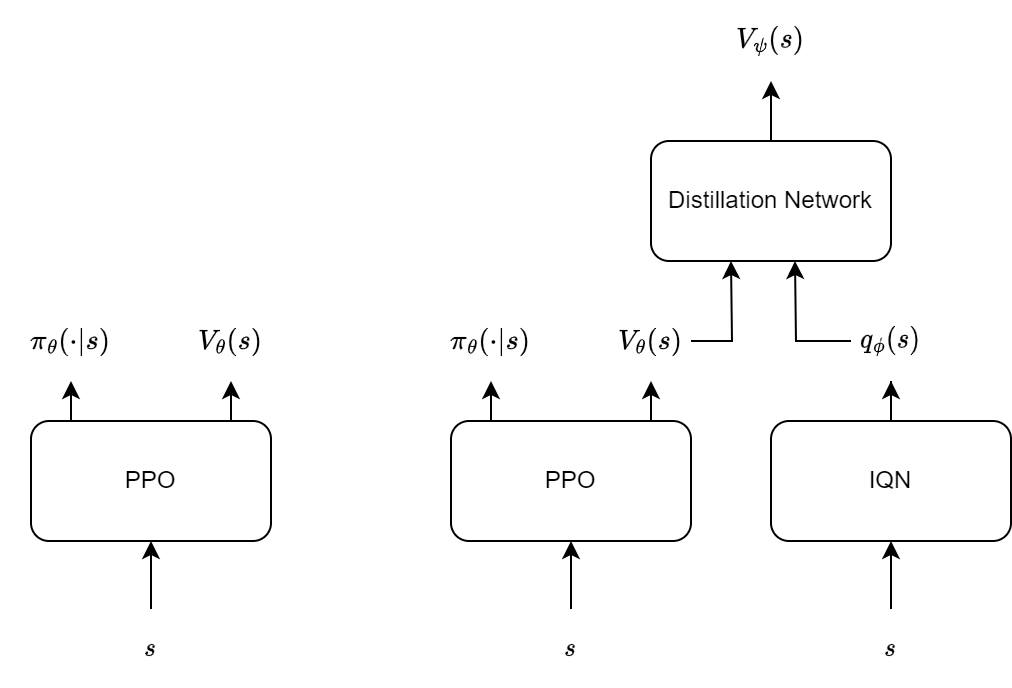}
  \caption{Comparison of model architecture between PPO and PG-Rainbow. PG-Rainbow uses a distillation network to incorporate quantile value distribution data into the policy network.}
  \label{fig:modelarch}
\end{figure}

\section{PG-Rainbow}
\subsection{Analyzing Multi-Modality of Rewards}
Though PPG and DCPG improve the vanilla PPO algorithm by tuning the value function, they still fall short in fully leveraging the distributional nature of the environment and the inherent uncertainty in reinforcement learning tasks. Realizing the distributional nature of returns in an environment is crucial for a better understanding of the environment, as the underlying reward distribution is rarely in the form of a normal distribution. In other words, it is important for the policy agent to take into account of the multi-modality of the reward distribution per action while seeking the optimal policy. However, current policy gradient methods fall short in incorporating the multi-modal distribution of the environment, as the advantage function used in the objective function is calculated with a single scalar value $V(s)$. 

\begin{figure}[H]
  \centering
  \includegraphics[scale=0.4]{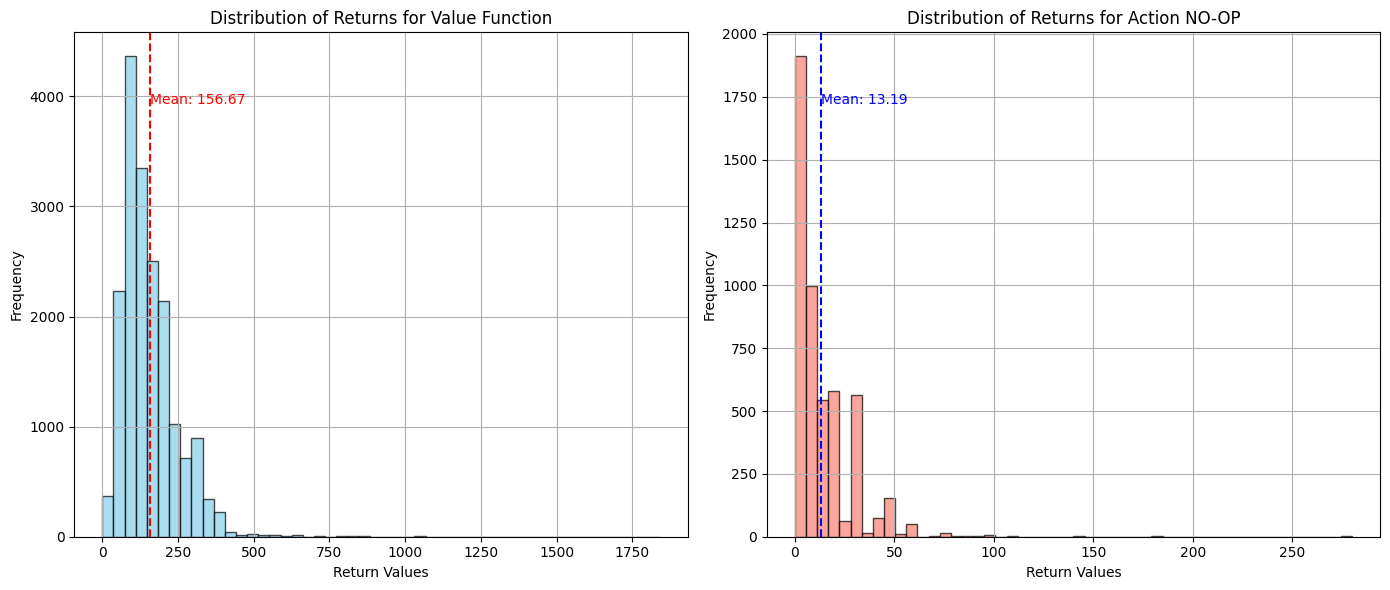}
  \caption{Histograms depicting the distribution of returns for the value function and action NO-OP in the DemonAttack environment of the Atari suite. The blue histogram represent the returns captured by the value function, while the orange histogram depict the returns captured by the Q function. The disparity between the distributions highlights the limitations of the value function in accurately capturing the range of returns for each action value.}
  \label{fig:return_dist}
\end{figure}

To show how current policy gradient methods lack in capturing the reward distribution while training, we conduct a simple experiment where we sample 100,000 trajectories from a PPO agent and plot the distribution of $V(s_0)$ values. We also sample 25,000 trajectories where $a_0$ is fixed to NO-OP in order to plot the distribution of $Q(s,a_0)$ values. Figure \ref{fig:return_dist} illustrates the results of this experiment.
The results demonstrate a notable discrepancy between the distribution of the value function and the Q-function for Action NO-OP. The value function's distribution is relatively centralized with a peak around moderate values concentrated around 100-150, indicating a general expectation of moderate returns across states. In contrast, the Q-function for Action NO-OP shows a highly skewed distribution towards lower values, suggesting that this action yields poor returns in most states. This divergence highlights a critical insight that relying solely on the value function can mask the variability and suboptimal performance of certain actions, potentially leading to suboptimal policy decisions. Therefore, it is imperative to consider the distributions of rewards when evaluating and improving policies. By incorporating these distributions, we can capture the nuances of different actions' performance across states, thereby enabling more robust and effective reinforcement learning models that avoid the pitfalls of average-based evaluations. This approach can lead to more accurate and optimal decision-making processes by acknowledging and addressing the inherent variability in action outcomes.

\subsection{Utilizing a distillation network}
Knowledge distillation is the process of transferring knowledge of a large teacher model to a relatively smaller child model. The transfer of knowledge may take different forms, from directly tuning the child model's network parameters to match the teacher's network parameters, to training the child model such that the model output matches that of the teacher's (\citet{knowledgedistillation}). Though PG-Rainbow is not a conventional application of knowledge distillation, we use the intuition behind the method such that the distributional reinforcement model transfers the value distribution information to the policy gradient model.

We propose that by incorporating an additional neural network, termed the distillation network, relevant value distribution information can be integrated into the value network of the policy gradient model, thereby enhancing the model's ability to capture the distributional characteristics of the environment. Employing a separate neural network for the distillation process is crucial, as the objective of the distillation network is to formulate a new value function that leverages information from both the policy gradient method and a distributional reinforcement learning method. This approach differs from model ensemble methods where the outputs of distinct models are merged to produce the final output.

\begin{figure}[H]
  \centering
  \includegraphics[scale=0.5]{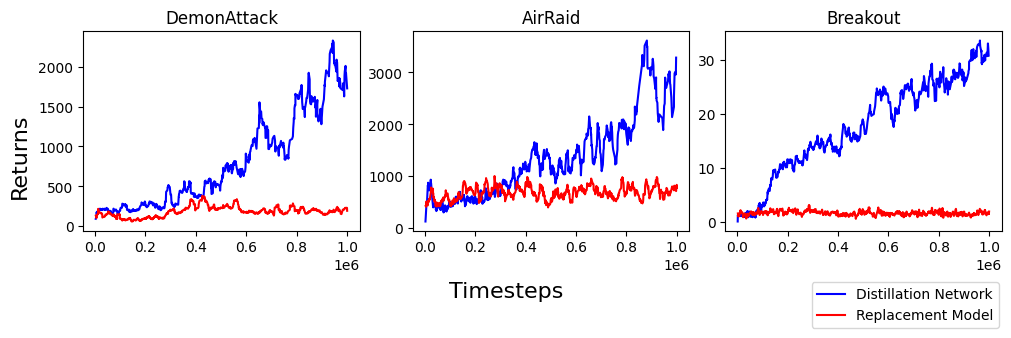}
  \caption{Comparison of performance between using a distillation network and replacing the value function. The use of a distillation network significantly outperforms a model where the PPO value function is replaced with the IQN value function. It is observable that the model with the replacement method completely fails to improve its performance.}
  \label{fig:replace}
\end{figure}

Figure \ref{fig:replace} illustrates the performance of different approaches for integrating a distributional reinforcement learning model with a policy gradient model in the Atari environment. In this experiment, we use the IQN as the distributional reinforcement learning model and PPO as the policy gradient model. We specifically compare the performance of using a distillation network with a model where the value function of the PPO is replaced by the value function of the IQN. This makes the model have disjoint policy and value networks, with entirely separate optimization processes. The results clearly demonstrate that employing disjoint networks without any representation sharing between the policy and value network leads to a degradation in the performance of the model, despite the value function capturing a more versatile understanding of the environment by being trained with a distributional reinforcement learning approach.

The necessity of using a distillation network rather than simply replacing the PPO value function with the IQN value function lies in the ability of the distillation network to integrate and harmonize the diverse types of information provided by both models. One might argue that, given the IQN model's capability to capture the rich dynamics of the environment's distributional nature, using the IQN value function with PPO is enough for the model to achieve enhanced performance. However as our empirical results show, a direct replacement can disrupt the balance and synergy between the policy and value networks in PPO, potentially leading to suboptimal learning dynamics and performance. In contrast, the distillation network allows for a smooth transfer of distributional value information from IQN to PPO, ensuring that the policy gradient model can leverage the detailed value distribution insights while maintaining the structural integrity and learning dynamics of the original PPO model. This hybrid approach ensures that the enhanced value information enriches the learning process, leading to superior overall performance in reinforcement learning tasks.

\subsection{Algorithm}
To address the issue of potential suboptimality arising from relying solely on the value function, we propose a novel method called PG-Rainbow that leverages a distillation network to incorporate quantile distribution information from the Implicit Quantile Network (IQN) into the Proximal Policy Optimization (PPO) value function head. This method enhances the value function by integrating the rich distributional information provided by IQN, capturing the variability and uncertainty in the expected returns for different actions. Consequently, the value function reflects not just the mean expected return, but also the spread and quantiles of the return distribution, offering a comprehensive understanding of potential outcomes and addressing issues highlighted in Section 2.2. By distilling quantile-based insights from IQN into the PPO framework, PG-Rainbow aims to improve the accuracy and robustness of policy evaluations, aligning learned policies more closely with true reward distributions. This approach mitigates the risk of suboptimal policy decisions arising from average-based evaluations and leverages the strengths of both distributional reinforcement learning and policy optimization techniques. We anticipate that PG-Rainbow will significantly enhance both the robustness and performance of agents across various tasks, demonstrating the efficacy of integrating distributional reinforcement learning with policy gradient methods to provide a nuanced and detailed representation of the value landscape.

Unlike common PPO algorithms that discard the experiences gathered by the agent after a policy update, PG-Rainbow stores them in a replay buffer $\mathcal{D}$. The stored experiences are used to train the IQN network $\phi$, where the IQN network provides the quantile values, later to be used in distilling the value distribution information to the critic network of PPO, parameterized with $\theta$. The distillation process is done by first taking the Hadamard product between the critic network output and the quantile values, which is fed into a distillation network $f_{\psi}$ composed of linear layers with a ReLU activation function (\citet{relu}). Hence, the final value of the critic network in PPO denoted $V_{\psi}(s)$ can be expressed as the following: \[
V_{\psi}(s) = f_{\psi}(V_{\theta}(s) \odot q_{\phi}(s))
\]
By adding value distribution information to the critic network, the policy agent becomes better equipped to assess the full spectrum of potential actions' consequences in a given state, thus enabling more informed decision-making processes within the reinforcement learning framework. Algorithm \ref{alg:algorithm} shows the process of training PG-Rainbow and Figure \ref{fig:modelarch} shows the architecture of PG-Rainbow compared to PPO.
\begin{algorithm}[H]
\caption{PG-Rainbow Pseudocode}
\label{alg:algorithm}
\hspace*{\algorithmicindent} \textbf{Input}: PPO network parameter $\theta$, IQN network parameter $\phi$, Replay buffer $\mathcal{D}$
\begin{algorithmic}[1]
\For{$\text{iteration}=1,2,\cdots$}
    \For{$\text{actor}=1,\cdots,N$}
        \State Run policy $\pi_{\theta_\text{old}}$ in environment for $T$ timesteps and store experiences in $\mathcal{D}$
        \State Train the IQN network using sampled experiences from $\mathcal{D}$
        \State Compute $\hat{A}_{t}$ for $t=1,\cdots,T$ in GAE style
    \EndFor
    \For{$\text{iteration}=1,\cdots,N$}
        \State Optimize loss function $L$ wrt $\theta$, with $K$ epochs and minibatch size $M$
        \State $\theta_\text{old}\leftarrow\theta$
    \EndFor
\EndFor
\end{algorithmic}
\end{algorithm}

\section{Experiments and Results}
\subsection{Implementation Details}
We evaluate the performance of PG-Rainbow in the Atari-2600 game suite, simulated via the Arcade Learning Environment (ALE) (\citet{atariale}) and follow the practice introduced by \citet{atariprprocessing} for preprocessing the game environment. We train our agent for 1 million frames and measure the average and standard deviation of returns. The performance is compared with the vanilla PPO algorithm. To ensure faster training time, we use an A2C-style (\citet{a2c}) approach to initialize 8 environments in parallel to collect experiences. We use a batch size of 32 to train the models. Note that hyperparameters and model architecture shared between PPO and PG-Rainbow, such as batch size or policy update epochs, are identical. The list of key hyperparameters used in this experiment is shown in Table \ref{tab:hyperparam}.
\begin{table}[]
\centering
\caption{Key hyperparameters used in experiments for PG-Rainbow}
\begin{tabular}{|ll|}
\hline
\multicolumn{2}{|c|}{Hyperparameters}                   \\ \hline
\multicolumn{1}{|l|}{Timesteps}               & 1000000 \\ \hline
\multicolumn{1}{|l|}{Number of environments}  & 8       \\ \hline
\multicolumn{1}{|l|}{Number of rollout steps} & 128     \\ \hline
\multicolumn{1}{|l|}{Batch Size}              & 32      \\ \hline
\multicolumn{1}{|l|}{Number of IQN quantiles} & 32      \\ \hline
\multicolumn{1}{|l|}{Number of update epochs} & 4       \\ \hline
\multicolumn{1}{|l|}{Learning rate}           & 2.5e-4  \\ \hline
\end{tabular}
\label{tab:hyperparam}
\end{table}

\subsection{Results}
\begin{figure}[h]
  \centering
  \includegraphics[scale=0.5]{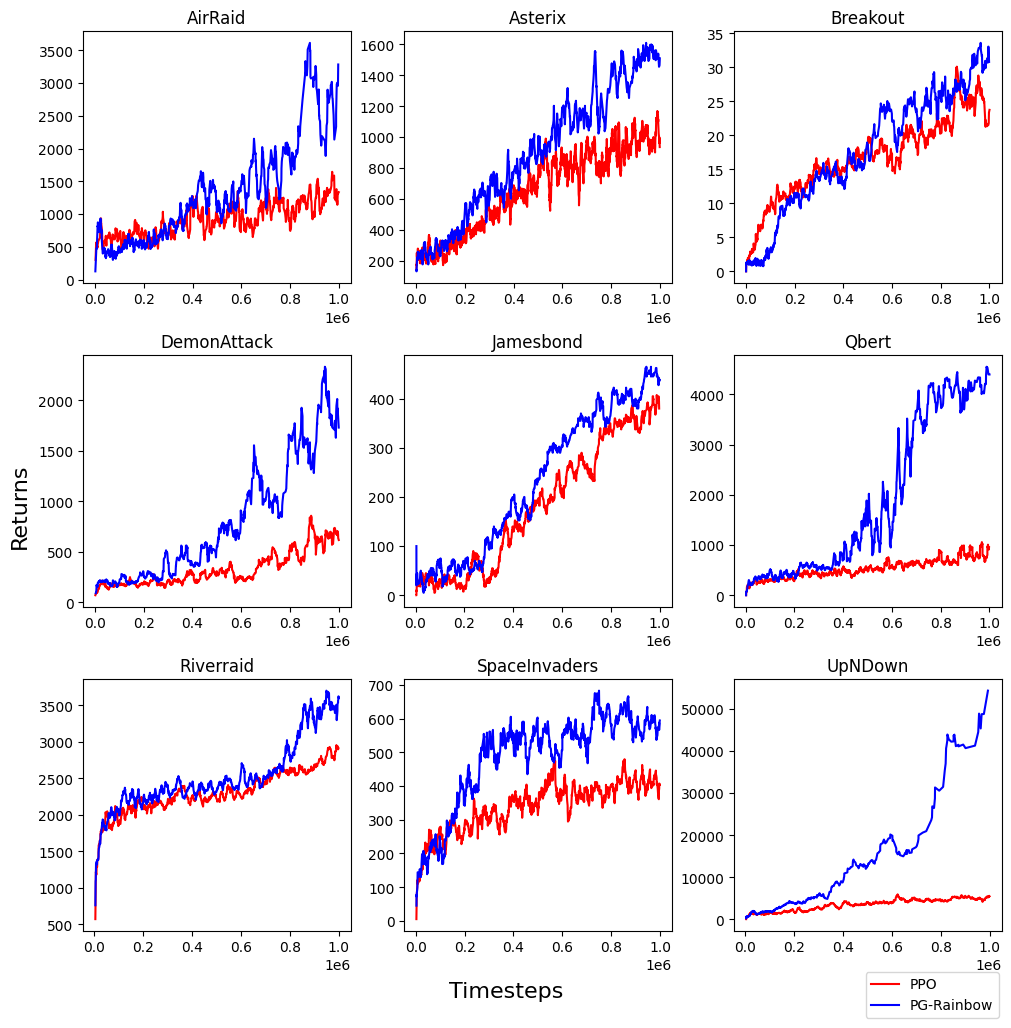}
  \caption{Performance of PG-Rainbow and PPO on Atari environments. Plots show the average episodic returns of the agent trained for 1 million timesteps. PG-Rainbow outperforms PPO in most environments.}
  \label{fig:pgr_vs_ppo}
\end{figure}
\begin{figure}[h]
  \centering
  \includegraphics[scale=0.5]{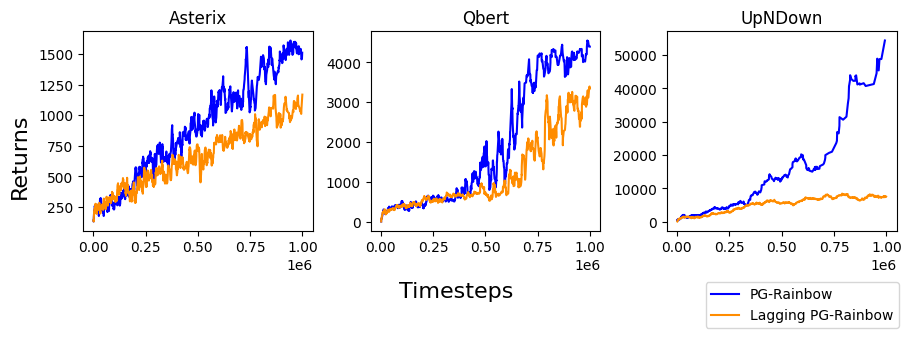}
  \caption{Performance of PG-Rainbow and Delayed PG-Rainbow on DemonAttack and BeamRider. Plots show the average episodic returns of the agent trained for 1 million timesteps. Results show that deferring the distillation process to the PPO agent adversely affects its performance.}
  \label{fig:delay_pgr}
\end{figure}
\begin{figure}[h]
  \centering
  \includegraphics[scale=0.4]{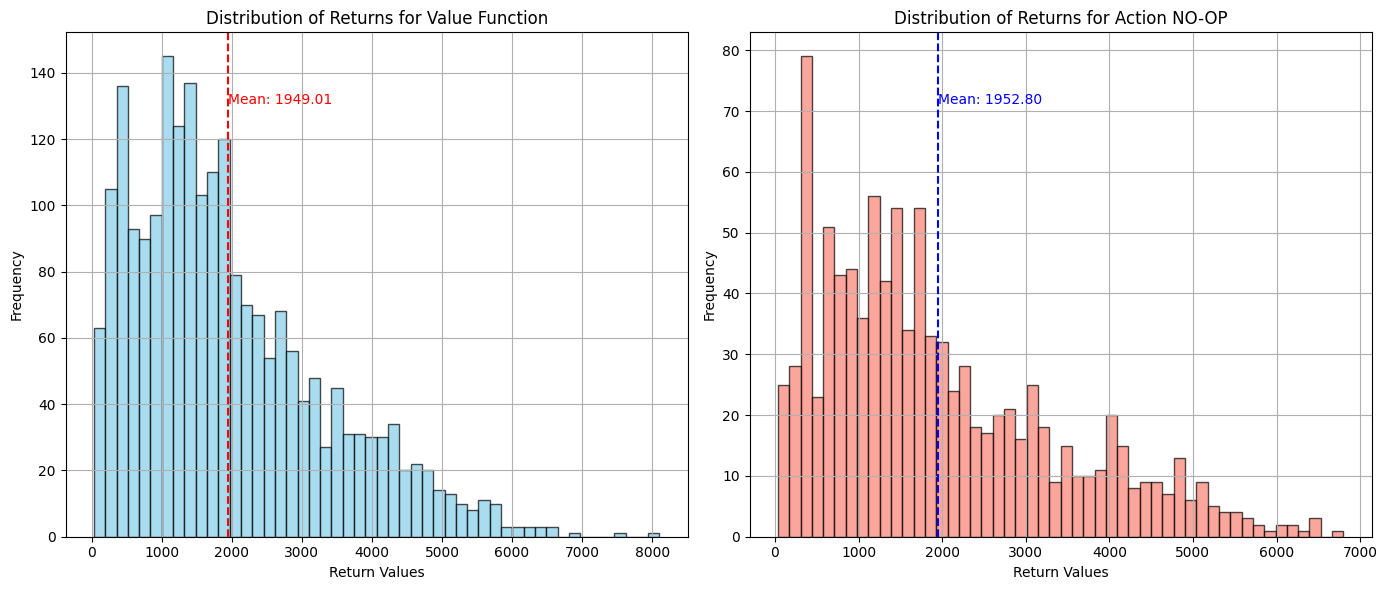}
  \caption{Histograms depicting the distribution of returns for the value function and action NO-OP in the DemonAttack environment of the Atari suite. The blue histogram represent the returns captured by the value function, while the orange histogram depict the returns captured by the Q Function.}
  \label{fig:pg-rainbow_dist}
\end{figure}
\begin{figure}[h]
  \centering
  \includegraphics[scale=0.5]{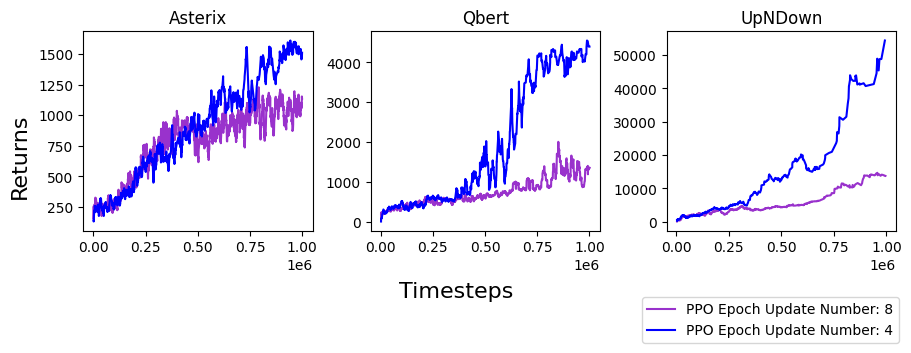}
  \caption{Performance of PG-Rainbow across different number of update epochs in PPO on DemonAttack and BeamRider. Plots show the average episodic returns of the agent trained for 1 million timesteps. Reducing the number of epochs show higher performance for PG-Rainbow.}
  \label{fig:epoch_change}
\end{figure}

We first compare the performance of PG-Rainbow with the PPO algorithm, shown in Figure \ref{fig:pgr_vs_ppo}. PG-Rainbow outperforms the PPO algorithm in most environments in terms of episodic returns for Atari environments. To prove the efficacy of integrating quantile information into the PPO algorithm, we undertake a series of experiments to elucidate the influence of adjustments made to the distillation network or the distributional reinforcement learning aspect on the performance of our model. First, we compare our PG-Rainbow algorithm with a delayed version, in which the quantile information is distilled to the critic network only after the 500,000th timestep. Note that delaying the distillation process does not mean that the IQN model isn't trained, but that the distributional information is not fed forward into the distillation network $\psi$. Figure \ref{fig:delay_pgr} shows the results of this experiment, where it is shown that when the quantile information is distilled to the PPO agent in a delayed manner, there is a significant degradation in performance. Next, we conduct the same experiment as in Section 3.1, where we plot the reward distribution of PG-Rainbow. We sample trajectories from the DemonAttack environment with the trained PG-Rainbow agent. The distribution of returns between the $V(s_0)$ values and $Q(s,a_0)$, where $a_0$ is the action NO-OP, is shown in Figure \ref{fig:pg-rainbow_dist}. Notably, the value function now captures the distributions of returns across actions more accurately than in our previous analysis of the regular PPO. The skewness in the return distributions is more accurately reflected in the value function of PG-Rainbow and it even captures the spike in return values within the 200-300 range. This is a significant improvement over regular PPO, where the value function primarily captures average expected returns without adequately reflecting the spread and variability of action-specific returns. The enhanced fidelity in capturing return distributions demonstrates the efficacy of incorporating quantile information through the distillation network. Consequently, we postulate that this integration leads to better-informed policy decisions and improved overall performance of the model.
We also experiment how the change in number of epochs for the PPO update within PG-Rainbow affects training, with all other hyperparameters equal. The results are shown in Figure \ref{fig:epoch_change}. It is noteworthy that diminishing the number of epochs allocat
ed for PPO training yields an enhancement in the algorithm's performance. We postulate that training with more epochs and using more samples does not necessarily improve the performance of the policy gradient algorithm. Instead of solely focusing on increasing the number of epochs, our findings suggest that the pivotal factor for enhancing the performance of an actor-critic style policy gradient lies in training a more accurate value function. This is achieved through the utilization of a distillation network in the case of PG-Rainbow. This assertion is consistent with the observations made by \citet{phasicpolicygradient}, wherein it was demonstrated that PPO derives advantages from increased sample reuse primarily due to the supplementary epochs facilitating additional training of the value function. 

\begin{figure}[h]
  \centering
  \includegraphics[scale=0.5]{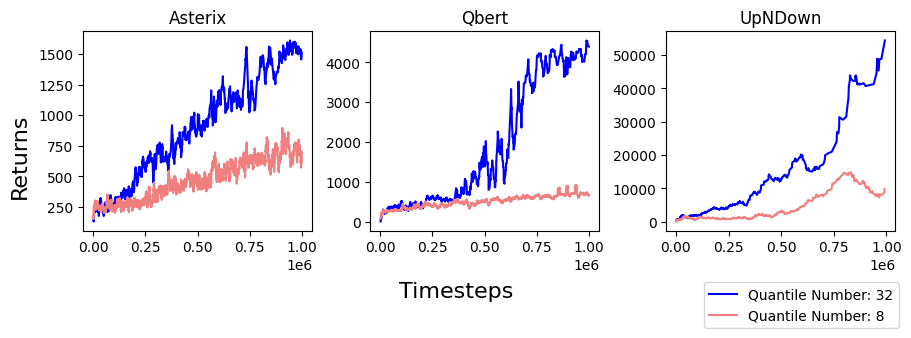}
  \caption{Performance of PG-Rainbow across different number of quantiles on DemonAttack and BeamRider. Plots show the average episodic returns of the agent trained for 1 million timesteps. A higher number of quantiles during training correlates with improved performance.}
  \label{fig:quantile_change}
\end{figure}

To further support this claim, we train PG-Rainbow by varying the number of quantiles the IQN model provides to the distillation network. The results are shown in Figure \ref{fig:quantile_change}. It was shown that increasing the number of quantiles enables the IQN model to capture a more comprehensive representation of the distribution of rewards, thereby resulting in improved performance (\citet{iqn}). Our results corroborates this finding as well, as the performance of PG-Rainbow increases as the number of quantiles provided to the distillation network is increased.
\begin{figure}[H]
  \centering
  \includegraphics[scale=0.5]{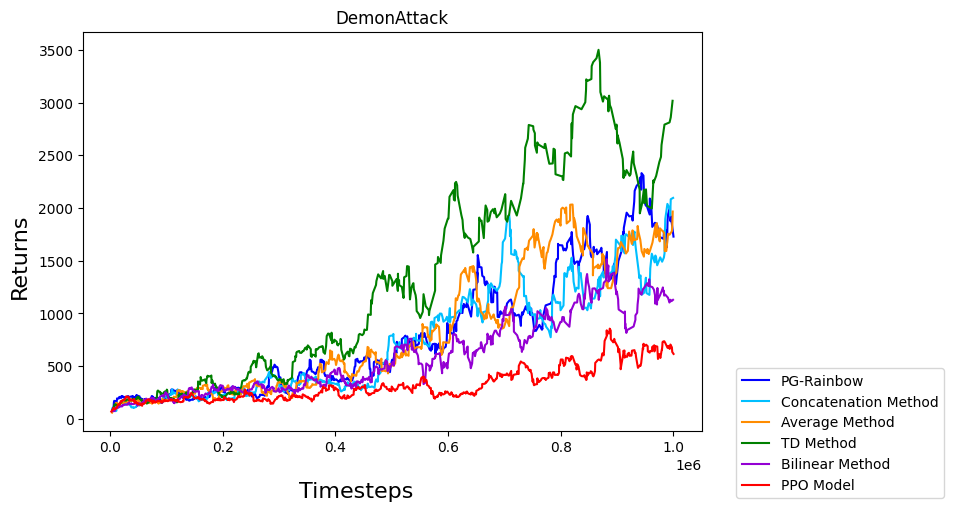}
  \caption{Performance of PG-Rainbow with different methods of computing input to the distillation network. All methods surpass the vanilla PPO's performance.}
  \label{fig:method_comparison}
\end{figure}
As introduced in Section 3.3, the input to the distillation network has been implemented as a pointwise multiplication of the scalar value function output from PPO and the quantile values from IQN. However, it is important to note that this is not the only viable method for integrating these inputs. The distillation network's input can be adapted in various forms, provided it allows the network to learn the relationship between the value function output and the quantile values effectively. The critical requirement for any input form is its structural form such that the distillation network can uncover and leverage the intricate relationships between the PPO value function and the distributional outputs from IQN. We test the following different methods of computing inputs to the distillation network.
\begin{enumerate}
  \item Concatenate the critic output from the PPO model with the quantile values.
  \[\text{Concat}(V_{\theta}(s), q_{\phi}(s))\]
  \item Average the critic output from the PPO model with the quantile values.
  \[\frac{1}{2}\cdot(V_{\theta}(s) + \frac{1}{32}\sum_{i=1}^{32} q_{\phi,i}(s))\]
  \item Compute the difference between the critic output and the mean of quantile values weighted by each action probabilities. 
  \[\pi_{\theta}(\cdot\lvert s)\times q_{\phi}(s)-V_{\theta}(s)\]
  \item Feed the critic output and quatile values to a Bilinear layer before the distillation network.
  \[\text{Bilinear}(V_{\theta}(s), q_{\phi}(s))\]
\end{enumerate}
Note that for method 3, we add the critic value function output, $V_{\theta}(s)$, with the distillation network output to derive the final value function output. The purpose of the distillation network in method 3 is to compute the difference between the quantile values scaled by the action-value probabilities provided the PPO model with the value function output of the PPO model. The difference is then added to the critic value function output, hence adjusting the value estimates based on the distributional information from IQN. The results are shown in Figure \ref{fig:method_comparison}. All the methods exceed the performance of the PPO model, with some methods even surpassing the performance of the originally proposed PG-Rainbow. However, our experiments show that the originally proposed PG-Rainbow method ensures the best stability, as other methods tend to have varying performance across environments. By exploring and optimizing these various input forms in future research, the integration process can be made more robust and effective, potentially leading to improved performance and stability.

\section{Conclusion}
In this work we proposed a novel method to incorporate distributional reinforcement learning with policy gradient methods. We have shown that by providing the value distribution information to the critic network of the policy gradient algorithm, the agent can make better informed actions, leading to higher episodic returns. By storing experiences gathered by the PPO network, we can utilize them to train the IQN network, alleviating the sample inefficiency problem of policy gradient methods. It is also worthwhile to note that our method PG-Rainbow incorporates both off-policy algorithms with on-policy reinforcement learning algorithms. To the best of our knowledge, few attempts have been made to formulate a reinforcement learning model where the strengths of both fields are combined. We hope that our work provides the foundation for further research in incorporating complementary methodologies from both off-policy and on-policy reinforcement learning paradigms, fostering the development of more robust and versatile approaches to solving complex decision-making problems.

Our work can be further improved and pave way for further research in the following domains. First, while using the IQN network represents a significant advancement, its inherent limitation lies in its applicability only to discrete action spaces. However, many real-world scenarios in fields such as robotics, healthcare, and finance demand efficient algorithms capable of operating in continuous action spaces. Thus, future research efforts could focus on extending and adapting the principles of implicit quantile networks to address this crucial requirement, enhancing their applicability and effectiveness in a broader range of domains. Second, our model architecture can be further fine-tuned to improve the efficacy of our algorithm. As discussed, DCPG uses a delayed optimization process to the value network with a single shared network architecture, thereby becoming a more light-weight and computationally efficient model. While PG-Rainbow yields improved episodic returns, it remains a computationally intensive model that necessitates longer training time compared to the vanilla PPO algorithm. Thus, future endeavors could focus on optimizing the computational efficiency of PG-Rainbow without compromising its performance, thereby making it more practical and scalable for real-world applications.

\bibliography{references}

\begin{thebibliography}{24}
\providecommand{\natexlab}[1]{#1}
\providecommand{\url}[1]{\texttt{#1}}
\expandafter\ifx\csname urlstyle\endcsname\relax
  \providecommand{\doi}[1]{doi: #1}\else
  \providecommand{\doi}{doi: \begingroup \urlstyle{rm}\Url}\fi

\bibitem[Agarap(2018)]{relu}
Abien~Fred Agarap.
\newblock Deep learning using rectified linear units (relu), 2018.
\newblock URL \url{http://arxiv.org/abs/1803.08375}.

\bibitem[Association and of~Mathematical~Statistics(1972)]{american1972annals}
American~Statistical Association and Institute of~Mathematical~Statistics.
\newblock \emph{The Annals of Mathematical Statistics}.
\newblock Number v. 43. Institute of Mathematical Statistics., 1972.
\newblock URL \url{https://books.google.co.kr/books?id=LE8KAAAAMAAJ}.

\bibitem[Bellemare et~al.(2012)Bellemare, Naddaf, Veness, and Bowling]{atariale}
Marc~G. Bellemare, Yavar Naddaf, Joel Veness, and Michael Bowling.
\newblock The arcade learning environment: An evaluation platform for general agents, 2012.
\newblock URL \url{http://arxiv.org/abs/1207.4708}.

\bibitem[Bellemare et~al.(2017)Bellemare, Dabney, and Munos]{DBLP:journals/corr/BellemareDM17}
Marc~G. Bellemare, Will Dabney, and R{\'{e}}mi Munos.
\newblock A distributional perspective on reinforcement learning, 2017.
\newblock URL \url{http://arxiv.org/abs/1707.06887}.

\bibitem[Bellemare et~al.(2023)Bellemare, Dabney, and Rowland]{distrlbook}
Marc~G. Bellemare, Will Dabney, and Mark Rowland.
\newblock \emph{Distributional Reinforcement Learning}.
\newblock MIT Press, 2023.
\newblock \url{http://www.distributional-rl.org}.

\bibitem[Bellman(1957)]{Bellman:DynamicProgramming}
Richard Bellman.
\newblock \emph{{Dynamic Programming}}.
\newblock Dover Publications, 1957.
\newblock ISBN 9780486428093.

\bibitem[Berner et~al.(2019)Berner, Brockman, Chan, Cheung, Debiak, Dennison, Farhi, Fischer, Hashme, Hesse, J{\'{o}}zefowicz, Gray, Olsson, Pachocki, Petrov, de~Oliveira~Pinto, Raiman, Salimans, Schlatter, Schneider, Sidor, Sutskever, Tang, Wolski, and Zhang]{DBLP:journals/corr/abs-1912-06680}
Christopher Berner, Greg Brockman, Brooke Chan, Vicki Cheung, Przemyslaw Debiak, Christy Dennison, David Farhi, Quirin Fischer, Shariq Hashme, Christopher Hesse, Rafal J{\'{o}}zefowicz, Scott Gray, Catherine Olsson, Jakub Pachocki, Michael Petrov, Henrique~Pond{\'{e}} de~Oliveira~Pinto, Jonathan Raiman, Tim Salimans, Jeremy Schlatter, Jonas Schneider, Szymon Sidor, Ilya Sutskever, Jie Tang, Filip Wolski, and Susan Zhang.
\newblock Dota 2 with large scale deep reinforcement learning, 2019.
\newblock URL \url{http://arxiv.org/abs/1912.06680}.

\bibitem[Cobbe et~al.(2020)Cobbe, Hilton, Klimov, and Schulman]{phasicpolicygradient}
Karl Cobbe, Jacob Hilton, Oleg Klimov, and John Schulman.
\newblock Phasic policy gradient, 2020.
\newblock URL \url{https://arxiv.org/abs/2009.04416}.

\bibitem[Dabney et~al.(2017)Dabney, Rowland, Bellemare, and Munos]{qrdqn}
Will Dabney, Mark Rowland, Marc~G. Bellemare, and R{\'{e}}mi Munos.
\newblock Distributional reinforcement learning with quantile regression, 2017.
\newblock URL \url{http://arxiv.org/abs/1710.10044}.

\bibitem[Dabney et~al.(2018)Dabney, Ostrovski, Silver, and Munos]{iqn}
Will Dabney, Georg Ostrovski, David Silver, and R{\'{e}}mi Munos.
\newblock Implicit quantile networks for distributional reinforcement learning, 2018.
\newblock URL \url{http://arxiv.org/abs/1806.06923}.

\bibitem[Haarnoja et~al.(2018)Haarnoja, Zhou, Abbeel, and Levine]{DBLP:journals/corr/abs-1801-01290}
Tuomas Haarnoja, Aurick Zhou, Pieter Abbeel, and Sergey Levine.
\newblock Soft actor-critic: Off-policy maximum entropy deep reinforcement learning with a stochastic actor, 2018.
\newblock URL \url{http://arxiv.org/abs/1801.01290}.

\bibitem[Hessel et~al.(2017)Hessel, Modayil, van Hasselt, Schaul, Ostrovski, Dabney, Horgan, Piot, Azar, and Silver]{rainbow}
Matteo Hessel, Joseph Modayil, Hado van Hasselt, Tom Schaul, Georg Ostrovski, Will Dabney, Daniel Horgan, Bilal Piot, Mohammad~Gheshlaghi Azar, and David Silver.
\newblock Rainbow: Combining improvements in deep reinforcement learning, 2017.
\newblock URL \url{http://arxiv.org/abs/1710.02298}.

\bibitem[Hinton et~al.(2015)Hinton, Vinyals, and Dean]{knowledgedistillation}
Geoffrey Hinton, Oriol Vinyals, and Jeff Dean.
\newblock Distilling the knowledge in a neural network, 2015.

\bibitem[Machado et~al.(2017)Machado, Bellemare, Talvitie, Veness, Hausknecht, and Bowling]{atariprprocessing}
Marlos~C. Machado, Marc~G. Bellemare, Erik Talvitie, Joel Veness, Matthew~J. Hausknecht, and Michael Bowling.
\newblock Revisiting the arcade learning environment: Evaluation protocols and open problems for general agents, 2017.
\newblock URL \url{http://arxiv.org/abs/1709.06009}.

\bibitem[Mnih et~al.(2013)Mnih, Kavukcuoglu, Silver, Graves, Antonoglou, Wierstra, and Riedmiller]{dqnatari}
Volodymyr Mnih, Koray Kavukcuoglu, David Silver, Alex Graves, Ioannis Antonoglou, Daan Wierstra, and Martin~A. Riedmiller.
\newblock Playing atari with deep reinforcement learning, 2013.
\newblock URL \url{http://arxiv.org/abs/1312.5602}.

\bibitem[Mnih et~al.(2016)Mnih, Badia, Mirza, Graves, Lillicrap, Harley, Silver, and Kavukcuoglu]{a2c}
Volodymyr Mnih, Adri{\`{a}}~Puigdom{\`{e}}nech Badia, Mehdi Mirza, Alex Graves, Timothy~P. Lillicrap, Tim Harley, David Silver, and Koray Kavukcuoglu.
\newblock Asynchronous methods for deep reinforcement learning, 2016.
\newblock URL \url{http://arxiv.org/abs/1602.01783}.

\bibitem[Moon et~al.(2023)Moon, Lee, and Song]{moon2023rethinking}
Seungyong Moon, JunYeong Lee, and Hyun~Oh Song.
\newblock Rethinking value function learning for generalization in reinforcement learning, 2023.

\bibitem[Raileanu and Fergus(2021)]{DBLP:journals/corr/abs-2102-10330}
Roberta Raileanu and Rob Fergus.
\newblock Decoupling value and policy for generalization in reinforcement learning, 2021.
\newblock URL \url{https://arxiv.org/abs/2102.10330}.

\bibitem[Schulman et~al.(2015)Schulman, Levine, Moritz, Jordan, and Abbeel]{DBLP:journals/corr/SchulmanLMJA15}
John Schulman, Sergey Levine, Philipp Moritz, Michael~I. Jordan, and Pieter Abbeel.
\newblock Trust region policy optimization, 2015.
\newblock URL \url{http://arxiv.org/abs/1502.05477}.

\bibitem[Schulman et~al.(2017)Schulman, Wolski, Dhariwal, Radford, and Klimov]{DBLP:journals/corr/SchulmanWDRK17}
John Schulman, Filip Wolski, Prafulla Dhariwal, Alec Radford, and Oleg Klimov.
\newblock Proximal policy optimization algorithms, 2017.
\newblock URL \url{http://arxiv.org/abs/1707.06347}.

\bibitem[Sutton and Barto(2018)]{Sutton1998}
Richard~S. Sutton and Andrew~G. Barto.
\newblock \emph{Reinforcement Learning: An Introduction}.
\newblock The MIT Press, second edition, 2018.
\newblock URL \url{http://incompleteideas.net/book/the-book-2nd.html}.

\bibitem[van Hasselt et~al.(2015)van Hasselt, Guez, and Silver]{doubledqn}
Hado van Hasselt, Arthur Guez, and David Silver.
\newblock Deep reinforcement learning with double q-learning, 2015.
\newblock URL \url{http://arxiv.org/abs/1509.06461}.

\bibitem[Wang et~al.(2015)Wang, de~Freitas, and Lanctot]{duelingdqn}
Ziyu Wang, Nando de~Freitas, and Marc Lanctot.
\newblock Dueling network architectures for deep reinforcement learning, 2015.
\newblock URL \url{http://arxiv.org/abs/1511.06581}.

\bibitem[Watkins(1989)]{Watkins}
C.~J. C.~H. Watkins.
\newblock Learning from delayed rewards, 1989.

\end{thebibliography}

\clearpage
\appendix
\section{Complete list of hyperparameters}
\begin{table}[htbp]
\centering
\label{tab:iqn_args}
\begin{tabular}{|l|p{0.5\linewidth}|l|}
\hline
\textbf{Argument} & \textbf{Description} & \textbf{Default Value} \\ \hline
total\_timesteps & Total timesteps of the experiments & 1000000 \\ \hline
learning\_rate & The learning rate of the optimizer & 2.5e-4 \\ \hline
num\_envs & The number of parallel game environments & 8 \\ \hline
num\_steps & The number of steps to run in each environment per policy rollout & 128 \\ \hline
anneal\_lr & Toggle learning rate annealing for policy and value networks & True \\ \hline
gamma & The discount factor gamma & 0.99 \\ \hline
gae\_lambda & The lambda for the general advantage estimation & 0.95 \\ \hline
num\_minibatches & The number of mini-batches & 4 \\ \hline
update\_epochs & The K epochs to update the policy & 4 \\ \hline
norm\_adv & Toggles advantages normalization & True \\ \hline
clip\_coef & The surrogate clipping coefficient & 0.1 \\ \hline
clip\_vloss & Toggles whether or not to use a clipped loss for the value function, as per the paper & True \\ \hline
ent\_coef & Coefficient of the entropy & 0.01 \\ \hline
vf\_coef & Coefficient of the value function & 0.5 \\ \hline
max\_grad\_norm & The maximum norm for the gradient clipping & 0.5 \\ \hline
target\_kl & The target KL divergence threshold & None \\ \hline
iqn\_start & Timestep to distill quantile information to PPO & 0 \\ \hline
num\_quantiles & Number of quantiles in IQN & 32 \\ \hline
batch\_size & The batch size (computed in runtime) & 1024 \\ \hline
minibatch\_size & The mini-batch size (computed in runtime) & 32 \\ \hline
num\_iterations & The number of iterations (computed in runtime) & 1953 \\ \hline
\end{tabular}
\end{table}
\clearpage

\section{Complete Atari Training Results (PG-Rainbow)}
\begin{figure}[H]
  \centering
  \includegraphics[width=\linewidth]{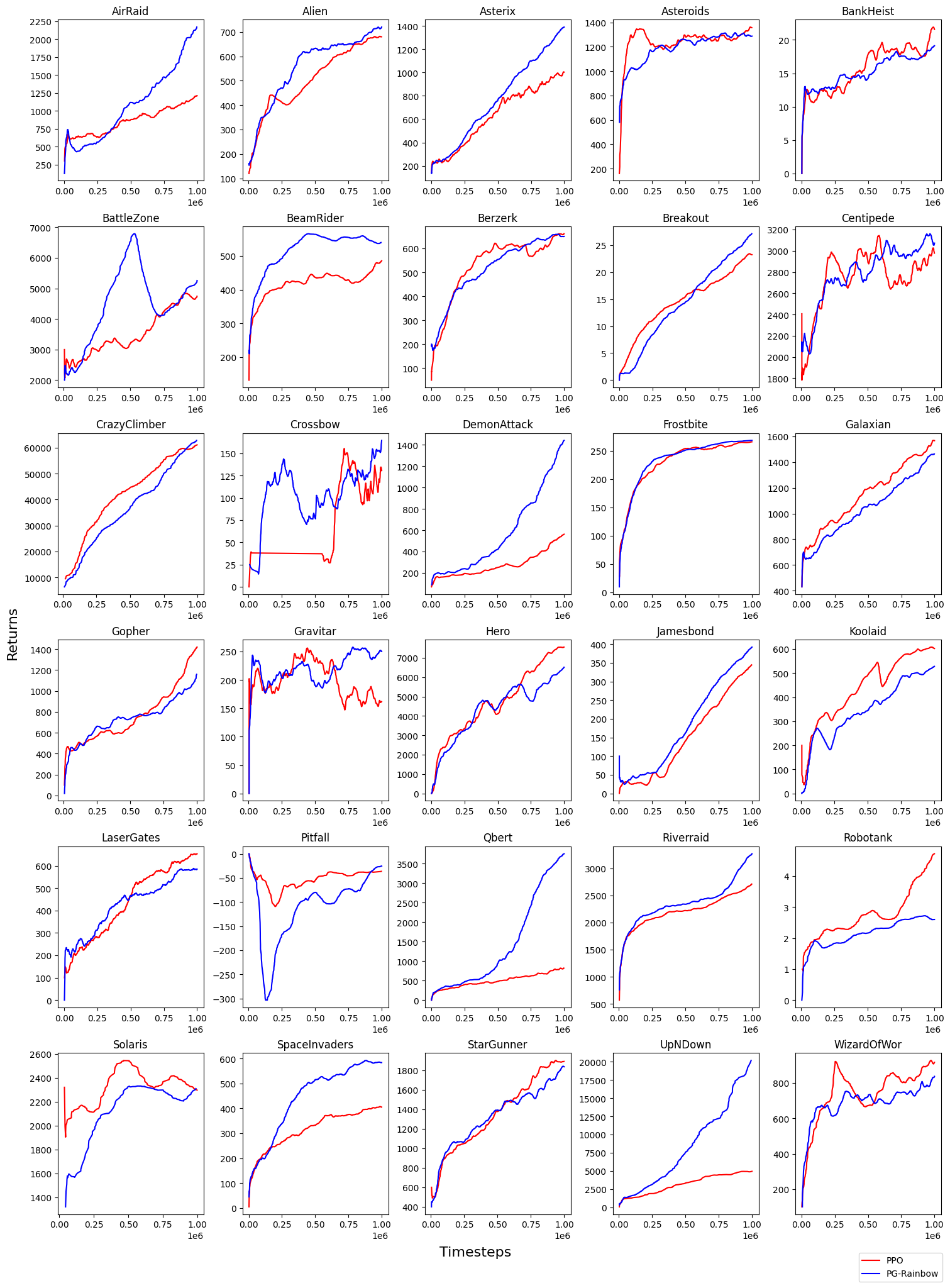}
\end{figure}
\clearpage

\begin{table}[h]
\centering
\caption{Comparison of raw scores between PG-Rainbow and PPO}
\begin{tabular}{lcc}
\toprule
Game & PG-Rainbow & PPO \\
\midrule
AirRaid & \textbf{3283.75} & 1333.75 \\
Alien & \textbf{803.0} & 714.0 \\
Asterix & \textbf{1510.0} & 935.0 \\
Asteroids & \textbf{1336.5} & 1332.5 \\
BankHeist & \textbf{18.5} & 18.5 \\
BattleZone & \textbf{9950.0} & 5650.0 \\
BeamRider & \textbf{585.20} & 517.60 \\
Berzerk & 642.5 & \textbf{697.5} \\
Breakout & \textbf{30.80} & 23.75 \\
Centipede & \textbf{3494.10} & 2548.55 \\
CrazyClimber & \textbf{74900.0} & 69935.0 \\
Crossbow & \textbf{267.5} & 90.0 \\
DemonAttack & \textbf{1728.75} & 616.5 \\
Frostbite & \textbf{269.0} & 267.5 \\
Galaxian & \textbf{1579.5} & 1566.5 \\
Gopher & \textbf{1761.0} & 1639.0 \\
Gravitar & \textbf{232.5} & 182.5 \\
Hero & \textbf{7914.75} & 7747.25 \\
Jamesbond & \textbf{437.5} & 380.0 \\
Koolaid & \textbf{540.0} & 570.0 \\
LaserGates & 650.75 & \textbf{701.25} \\
Pitfall & \textbf{-13.80} & -20.0 \\
Qbert & \textbf{4393.75} & 955.0 \\
Riverraid & \textbf{3599.0} & 2909.5 \\
Robotank & \textbf{2.75} & 7.0 \\
Solaris & \textbf{2636.0} & 2021.0 \\
SpaceInvaders & \textbf{594.5} & 403.5 \\
StarGunner & 1735.0 & \textbf{1940.0} \\
UpNDown & \textbf{54313.0} & 5503.5 \\
WizardOfWor & 970.0 & \textbf{1095.0} \\
\bottomrule
\end{tabular}
\end{table}

\clearpage
\section{Comparsion of results with Lagging PG-Rainbow}
\begin{figure}[H]
  \centering
  \includegraphics[width=\linewidth]{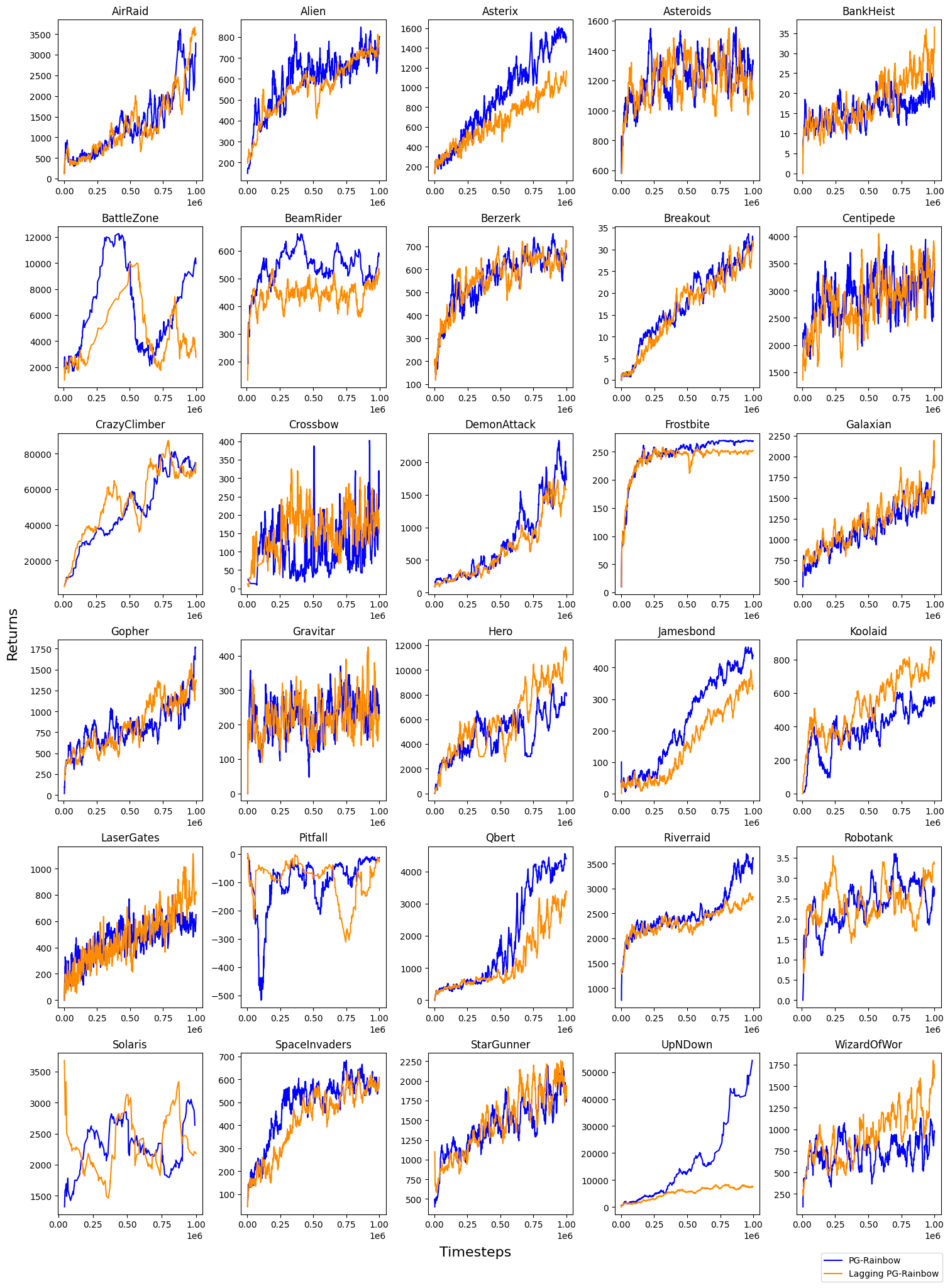}
\end{figure}
\clearpage

\begin{table}[h]
\centering
\caption{Comparison of raw scores between PG-Rainbow and Lagging PG-Rainbow}
\begin{tabular}{lcc}
\toprule
Game & PG-Rainbow & PPO \\
\midrule
AirRaid       & \bfseries 2711.25 & 1333.75 \\
Alien         & 629.0 & \bfseries 714.0 \\
Asterix       & \bfseries 1295.0 & 935.0 \\
Asteroids     & 1270.5 & \bfseries 1332.5 \\
BankHeist     & \bfseries 42.0 & 18.5 \\
BattleZone    & 3100.0 & \bfseries 5650.0 \\
BeamRider     & \bfseries 648.0 & 517.6 \\
Berzerk       & \bfseries 754.0 & 697.5 \\
Breakout      & \bfseries 26.65 & 23.75 \\
Centipede     & 2506.70 & \bfseries 2548.55 \\
CrazyClimber  & \bfseries 75735.0 & 69935.0 \\
Crossbow      & \bfseries 172.5 & 90.0 \\
DemonAttack   & \bfseries 3016.25 & 616.5 \\
Frostbite     & 257.0 & \bfseries 267.5 \\
Galaxian      & 1540.5 & \bfseries 1566.5 \\
Gopher        & 1319.0 & \bfseries 1639.0 \\
Gravitar      & \bfseries 232.5 & 182.5 \\
Hero          & 4614.5 & \bfseries 7747.25 \\
Jamesbond     & 307.5 & \bfseries 380.0 \\
Koolaid       & 510.0 & \bfseries 570.0 \\
LaserGates    & \bfseries 939.25 & 701.25 \\
Pitfall       & \bfseries -25.0 & -20.0 \\
Qbert         & \bfseries 3712.5 & 955.0 \\
Riverraid     & \bfseries 3054.5 & 2909.5 \\
Robotank      & 4.35 & \bfseries 7.0 \\
Solaris       & \bfseries 2024.0 & 2021.0 \\
SpaceInvaders & \bfseries 536.0 & 403.5 \\
StarGunner    & \bfseries 2200.0 & 1940.0 \\
UpNDown       & \bfseries 16456.0 & 5503.5 \\
WizardOfWor   & \bfseries 1165.0 & 1095.0 \\
\bottomrule
\end{tabular}
\end{table}

\clearpage
\section{Complete Atari Training Results (TD-Method)}
\begin{figure}[H]
  \centering
  \includegraphics[width=\linewidth]{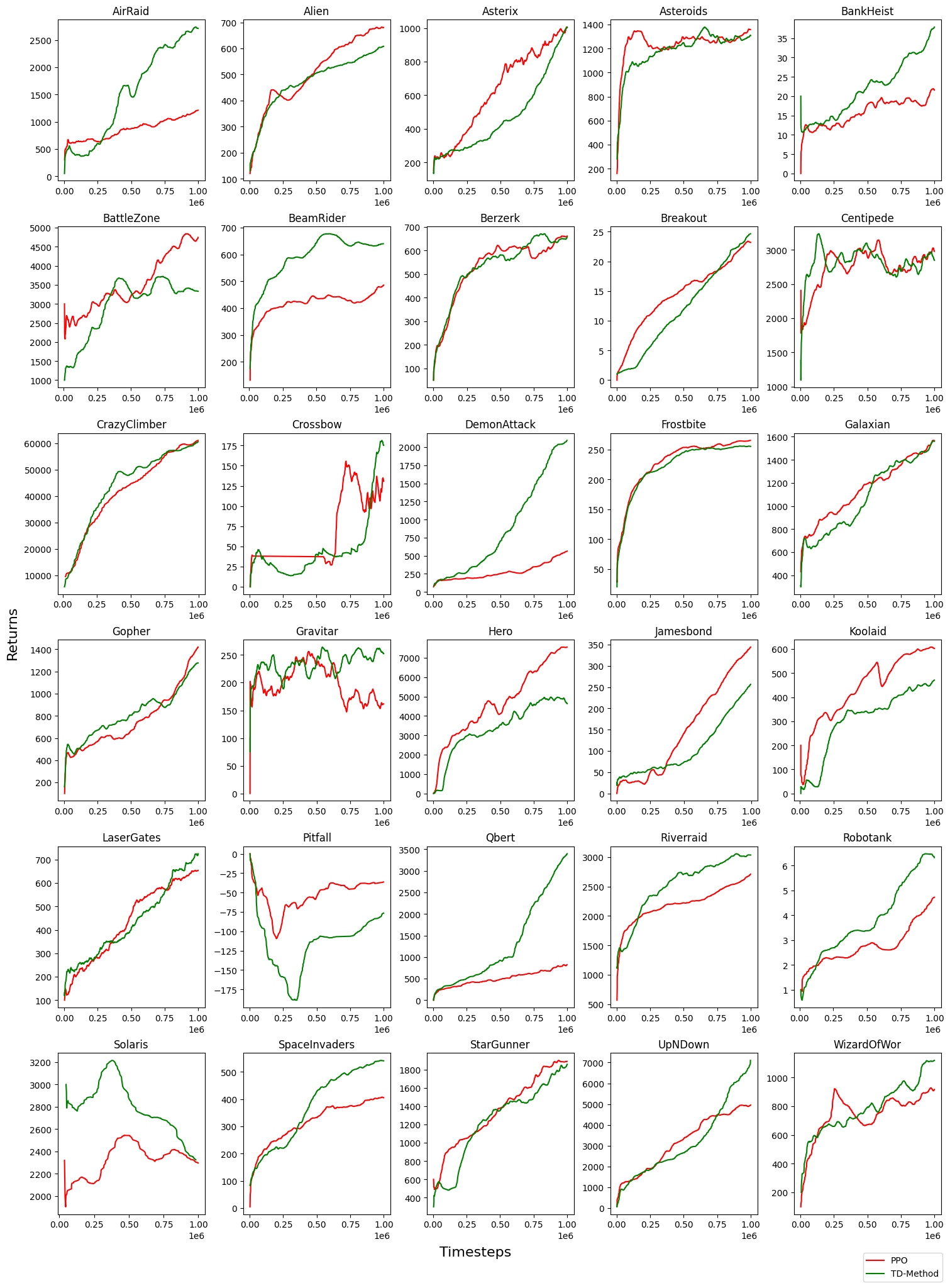}
\end{figure}

\clearpage
\begin{table}[h]
\centering
\caption{Comparison of raw scores between TD-Method and PPO}
\begin{tabular}{lcc}
\toprule
Game & TD-Method & PPO \\
\midrule
AirRaid       & \bfseries 2711.25 & 1333.75 \\
Alien         & 629.0 & \bfseries 714.0 \\
Asterix       & \bfseries 1295.0 & 935.0 \\
Asteroids     & 1270.5 & \bfseries 1332.5 \\
BankHeist     & \bfseries 42.0 & 18.5 \\
BattleZone    & 3100.0 & \bfseries 5650.0 \\
BeamRider     & \bfseries 648.0 & 517.6 \\
Berzerk       & \bfseries 754.0 & 697.5 \\
Breakout      & \bfseries 26.65 & 23.75 \\
Centipede     & 2506.70 & \bfseries 2548.55 \\
CrazyClimber  & \bfseries 75735.0 & 69935.0 \\
Crossbow      & \bfseries 172.5 & 90.0 \\
DemonAttack   & \bfseries 3016.25 & 616.5 \\
Frostbite     & 257.0 & \bfseries 267.5 \\
Galaxian      & 1540.5 & \bfseries 1566.5 \\
Gopher        & 1319.0 & \bfseries 1639.0 \\
Gravitar      & \bfseries 232.5 & 182.5 \\
Hero          & 4614.5 & \bfseries 7747.25 \\
Jamesbond     & 307.5 & \bfseries 380.0 \\
Koolaid       & 510.0 & \bfseries 570.0 \\
LaserGates    & \bfseries 939.25 & 701.25 \\
Pitfall       & \bfseries -25.0 & -20.0 \\
Qbert         & \bfseries 3712.5 & 955.0 \\
Riverraid     & \bfseries 3054.5 & 2909.5 \\
Robotank      & 4.35 & \bfseries 7.0 \\
Solaris       & \bfseries 2024.0 & 2021.0 \\
SpaceInvaders & \bfseries 536.0 & 403.5 \\
StarGunner    & \bfseries 2200.0 & 1940.0 \\
UpNDown       & \bfseries 16456.0 & 5503.5 \\
WizardOfWor   & \bfseries 1165.0 & 1095.0 \\
\bottomrule
\end{tabular}
\end{table}

\end{document}